\documentclass[letterpaper, 10pt, conference]{ieeeconf}

\IEEEoverridecommandlockouts                              % This command is only needed if
\usepackage{amsthm}
\usepackage{thmtools}
\declaretheoremstyle[%
  spaceabove=0pt,%reduce or increase between theorem and proof
  spacebelow=0pt,%reduce or increase
  headfont=\normalfont\itshape,%
  postheadspace=1em,%
  qed=\qedsymbol%
]{mystyle}
\declaretheorem[name={Proof},style=mystyle,unnumbered,
]{myproof}

\usepackage{amsmath, amssymb, amsfonts}
\usepackage[linesnumbered, ruled, vlined]{algorithm2e}
\SetArgSty{textrm}
\SetInd{0.1em}{0.5em}
\usepackage{algpseudocode}
\usepackage{array}
\usepackage{caption}
\usepackage{subcaption}
\usepackage{textcomp}
\usepackage{stfloats}
\usepackage{url}
\usepackage{verbatim}
\usepackage{graphicx}
\usepackage{cite}
\usepackage{multicol}
\usepackage{multirow}
\usepackage{soul}
\usepackage{array}
\usepackage{makecell}
\usepackage{xcolor}
\usepackage{cases}
\usepackage[hidelinks]{hyperref}

\let\labelindent\relax
\usepackage[shortlabels]{enumitem}

\DeclareMathOperator*{\argmin}{arg\,min\,}
\DeclareMathOperator*{\argmax}{arg\,max\,}
\DeclareMathOperator*{\minimize}{minimize}

\makeatletter
\def\thm@space@setup{%
 \thm@preskip=\parskip \thm@postskip=0pt
}
\makeatother

\SetCommentSty{mycommfont}

\newtheorem{theorem}{Theorem}
\renewcommand\qedsymbol{$\blacksquare$}
\newtheorem{lemma}[theorem]{Lemma}

\begin{document}

\title{\LARGE \bf Mixed Integer Programming for Time-Optimal Multi-Robot Coverage Path Planning with Efficient Heuristics}

\author{
Jingtao Tang$^{1}$, Hang Ma$^1$
% <-this % stops a space
\thanks{$^{1}$The authors are with the School of Computing Science, Simon Fraser University, Burnaby, BC V5A1S6, Canada {\tt\footnotesize \{jta204, hangma\}@sfu.ca}.}
% <-this % stops a space
% \thanks{$^\dagger$ The corresponding author.}
}

\markboth{Journal of \LaTeX\ Class Files,~Vol.~14, No.~8, August~2021}%
{Shell \MakeLowercase{\textit{et al.}}: A Sample Article Using IEEEtran.cls for IEEE Journals}



% \markboth{IEEE Robotics and Automation Letters. Preprint Version. Accepted August, 2023}
% {Tang \MakeLowercase{\textit{et al.}}: Mixed Integer Programming for Time-Optimal Multi-Robot Coverage Path Planning with Efficient Heuristics} 

\maketitle
\thispagestyle{empty}
\pagestyle{empty}

\begin{abstract}
We investigate time-optimal Multi-Robot Coverage Path Planning (MCPP) for both unweighted and weighted terrains, which aims to minimize the coverage time, defined as the maximum travel time of all robots. Specifically, we focus on a reduction from MCPP to Min-Max Rooted Tree Cover (MMRTC). For the first time, we propose a Mixed Integer Programming (MIP) model to optimally solve MMRTC, resulting in an MCPP solution with a coverage time that is provably at most four times the optimal.
Moreover, we propose two suboptimal yet effective heuristics that reduce the number of variables in the MIP model, thus improving its efficiency for large-scale MCPP instances. We show that both heuristics result in reduced-size MIP models that remain complete (i.e., guaranteed to find a solution if one exists) for all MMRTC instances.
Additionally, we explore the use of model optimization warm-startup to further improve the efficiency of both the original MIP model and the reduced-size MIP models. We validate the effectiveness of our MIP-based MCPP planner through experiments that compare it with two state-of-the-art MCPP planners on various instances, demonstrating a reduction in the coverage time by an average of $27.65\%$ and $23.24\%$ over them, respectively.
\end{abstract}

% \begin{IEEEkeywords}
% Path Planning for Multiple Mobile Robots or Agents; Multi-Robot Systems; Combinatorial Optimization
% \end{IEEEkeywords}

\section{Introduction}
Coverage Path Planning (CPP)~\cite{galceran2013survey} involves finding an optimal path for a robot to traverse a terrain of interest to completely cover all its regions. Examples of such terrains include indoor environments covered by vacuum cleaning robots~\cite{bormann2018indoor} and outdoor fields covered by unmanned aerial vehicles~\cite{lin2009uav}.
%In recent decades, the robotics community has given significant attention to CPP, resulting in a considerable amount of research focused on addressing this problem~\cite{galceran2013survey}.
Multi-Robot Coverage Path Planning (MCPP) is an extension of CPP that involves coordinating the paths of multiple robots to achieve complete coverage of the given terrain, thus improving coverage task efficiency and system robustness.
Therefore, MCPP plays a crucial role in various applications, including search and rescue~\cite{song2022multi}, environmental monitoring~\cite{collins2021scalable}, and mapping~\cite{azpurua2018multi}.
Developing efficient and robust MCPP algorithms is essential to enable the widespread deployment of multi-robot systems in robotics applications.

One of the main challenges in MCPP is to effectively distribute robots over the terrain to be covered while avoiding collisions with static obstacles.
This requires considering factors such as the traversability of the terrain, the mobility of the robots, and the availability of communication between the robots.
Additionally, the problem becomes increasingly complex as the number of robots increases, and we must consider more potential paths and interactions between robots.
Another challenge in MCPP is to ensure complete coverage, i.e., all regions of the terrain are covered, which can be challenging to achieve without sacrificing efficiency or incurring high computational costs.

This paper considers the fundamental challenge of time-optimal multi-robot planning in MCPP, which aims to minimize the coverage time, defined as the maximum travel time of all robots.
We focus on MCPP applications in agriculture and environmental monitoring, where path execution time is assumed to be orders of magnitude higher than the planning time, making solution quality more important than planning runtime. Offline planning is commonly used under this assumption to solve MCPP without inter-robot communication.
To solve MCPP, we adopt a problem formulation similar to that used in the existing literature, where the terrain to be covered is abstracted as a graph, and coverage time is evaluated as the sum of weights along the coverage paths on the graph. We conclude our main contributions as follows:
\begin{enumerate}
\item We propose a Mixed Integer Programming (MIP) model to optimally solve Min-Max Rooted Tree Cover (MMRTC), which results in an MCPP solution with an asymptotic optimality ratio of 4. We prove the correctness of the proposed MIP model.
\item Based on the proposed MIP model, we design two efficient suboptimal heuristics, the Parabolic Removal Heuristic and the Subgraph Removal Heuristic, which reduce the model size with a configurable loss of optimality. We prove that the two reduced-size models are complete (i.e., guarantee to find a solution if one exists) for all MMRTC instances.
\item We provide open-source code for our MIP-based MCPP planner and thorough experimental results, including model optimization warm-startup and performance comparisons against two state-of-the-art MCPP planners. Our MIP-based MCPP planner yields higher-quality solutions at the cost of longer runtime.
\end{enumerate}

\section{Related Work}

In this section, we survey related work on MCPP and the use of MIP for multi-robot planning.

% \noindent\textbf{Single- and Multi-Robot Coverage Path Planning:}
% The CPP problem is a well-studied single-robot planning problem in which a path is planned for a robot to completely cover a given terrain.
% We refer interested readers to the extensive surveys~\cite{choset2001coverage, galceran2013survey} on CPP.
% Most MCPP algorithms build upon existing CPP algorithms by first decomposing the terrain to be covered into multiple sub-areas with coverage paths and then assigning the paths to multiple robots. In general, both CPP and MCPP algorithms can be categorized as decomposition-based or graph-based methods.
% Decomposition-based methods~\cite{acar2002morse, rekleitis2008efficient, karapetyan2017efficient} first geometrically decompose the terrain and then generate back-and-forth coverage paths on each sub-area, which are straightforward but do not consider non-uniform traversal costs specified by weighted terrains as graph-based methods.
% This paper focuses on graph-based methods~\cite{gabriely2002spiral, hazon2005redundancy, kapoutsis2017darp} that operate on a graph representation of the terrain to be covered, which will be discussed in detail in the next paragraph.
% More recently, several learning-based planners~\cite{theile2020uav, lakshmanan2020complete, tang2022learning} have also been proposed to solve some application-specific coverage problems, where the uncertainties of robots and environment are a major consideration for the problem.

\noindent\textbf{Single- and Multi-Robot Coverage Path Planning:}
MCPP is a generalization of CPP, which involves planning paths for multiple robots to cover a given terrain cooperatively. We refer interested readers to the comprehensive surveys~\cite{choset2001coverage, galceran2013survey} on CPP and its extensions.
Most MCPP algorithms build upon existing CPP algorithms by partitioning the terrain into multiple regions with coverage paths and then assigning the paths to multiple robots. In general, MCPP algorithms can be categorized as decomposition-based or graph-based methods.
Decomposition-based methods~\cite{acar2002morse, rekleitis2008efficient, karapetyan2017efficient} first partition the terrain geometrically and then generate zigzag coverage paths within each region. While these methods are simple, they are not suitable for weighted terrains with non-uniform traversal costs and obstacle-rich terrains like mazes due to their reliance on geometric partitioning. 
%they cannot account for non-uniform traversal costs that may arise from weighted terrains and do not work well for obstacle-rich terrains such as mazes due to the geometrical partitioning.
This paper focuses on graph-based methods~\cite{gabriely2002spiral, hazon2005redundancy, kapoutsis2017darp} that operate on a graph representation of the terrain to be covered. Graph-based methods consider varying traversal costs and provide more flexibility. We will discuss them in more detail in the next paragraph.
In addition to traditional approaches, learning-based planners~\cite{theile2020uav, lakshmanan2020complete, tang2022learning} have been developed for application-specific coverage problems, where the uncertainties of robots and environment are a major consideration.

\noindent\textbf{Graph-Based Spanning Tree Coverage:}
One well-known method for solving graph-based CPP is Spanning Tree Coverage (STC)~\cite{gabriely2001spanning}, where the terrain to be covered is abstracted into uniformly sampled vertices, and robots are allowed to traverse along graph edges connecting adjacent vertices.
% By choosing the vertex sampling distance to ensure that every region can be swept by traversing between adjacent vertices, the coverage problem can be solved by covering all the vertices of the graph.
While CPP can be solved optimally in unweighted terrains~\cite{gabriely2001spanning} and near-optimally in weighted terrains~\cite{zheng2007robot} in polynomial time using STC~\cite{gabriely2001spanning}, graph-based MCPP with STC has been proved to be NP-hard~\cite{zheng2010multirobot}.
Thus, much MCPP research has focused on designing polynomial-time approximation algorithms.
Multi-Robot STC (MSTC)~\cite{hazon2005redundancy} splits the STC path into segments based on the initial location of each robot and then assigns the segments to robots. MSTC extensions have been developed to construct better spanning trees and STC paths~\cite{agmon2006constructing}, find balanced cut points on the STC path~\cite{tang2021mstc}, and support scenarios with fault tolerance~\cite{sun2021ft} and turning minimization~\cite{lu2022tmstc}.
Multi-Robot Forest Coverage (MFC) \cite{zheng2005multi, zheng2007robot, zheng2010multirobot} extends the Rooted Tree Cover (RTC) algorithm \cite{even2004min} to generate multiple rooted subtrees to cover all vertices of the input graph and then generate coverage paths using STC on each subtree.
A closely related problem is Multi-Traveling Salesmen Problem (mTSP)~\cite{cheikhrouhou2021comprehensive} that aims to find optimal routes for multiple salesmen who start and end at a city and visit all the given cities. However, most mTSP algorithms that can handle large-scale instances are designed for Euclidean spaces. We are unaware of any mTSP algorithms that are directly applicable to graph-based MCPP.
%A closely related problem to the graph-based MCPP problem is the multi-Traveling Salesmen Problem (mTSP)~\cite{cheikhrouhou2021comprehensive}, which plans for multiple salesmen starting and ending at a city to visit all the given cities. However, most mTSP algorithms that are scalable to large-size instances are designed for Euclidean spaces. We are not aware of any mTSP algorithms that are directly applicable to the graph-based MCPP problem. However, most mTSP algorithms are not scalable to MCPP since the problem size would be too large if we consider each vertex to be covered as the city to be visited. In addition, mTSP is defined on Euclidean space whereas graph-based MCPP problems have explicit problem structure on grid-like graphs.

\noindent\textbf{Integer Programming for Multi-Robot Planning:}
Integer Programming (IP) is a mathematical optimization technique used to solve problems that involve discrete variables under problem-specific constraints.
When the problem model also involves continuous variables, it is known as Mixed Integer Programming (MIP).
Both IP and MIP models can be solved to optimal via branch-and-bound methods on top of linear relaxation of the model, given enough time, and have been widely used in multi-robot planning.
Examples include a general IP framework~\cite{han2019integer} for Multi-Robot Path Planning and Multi-Robot Minimum Constraint Removal, a Branch-and-Cut-and-Price framework~\cite{lam2022branch} for Multi-Agent Path Finding, and a MIP model~\cite{lippi2021mixed} for Multi-Robot Task Allocation with humans.
Despite the effectiveness of MIP and IP in finding optimal solutions for various problems, as the problem size increases, the computational complexity of the MIP solver increases exponentially, making it challenging to solve large-scale instances in limited runtime.
Therefore, existing research has focused on designing efficient heuristics based on the MIP or IP framework~\cite{yu2016optimal, guo2021spatial}.

\section{Problem Formulation}\label{subsec:stc}

\begin{figure}[t]
\centering
\includegraphics[width=0.98\linewidth]{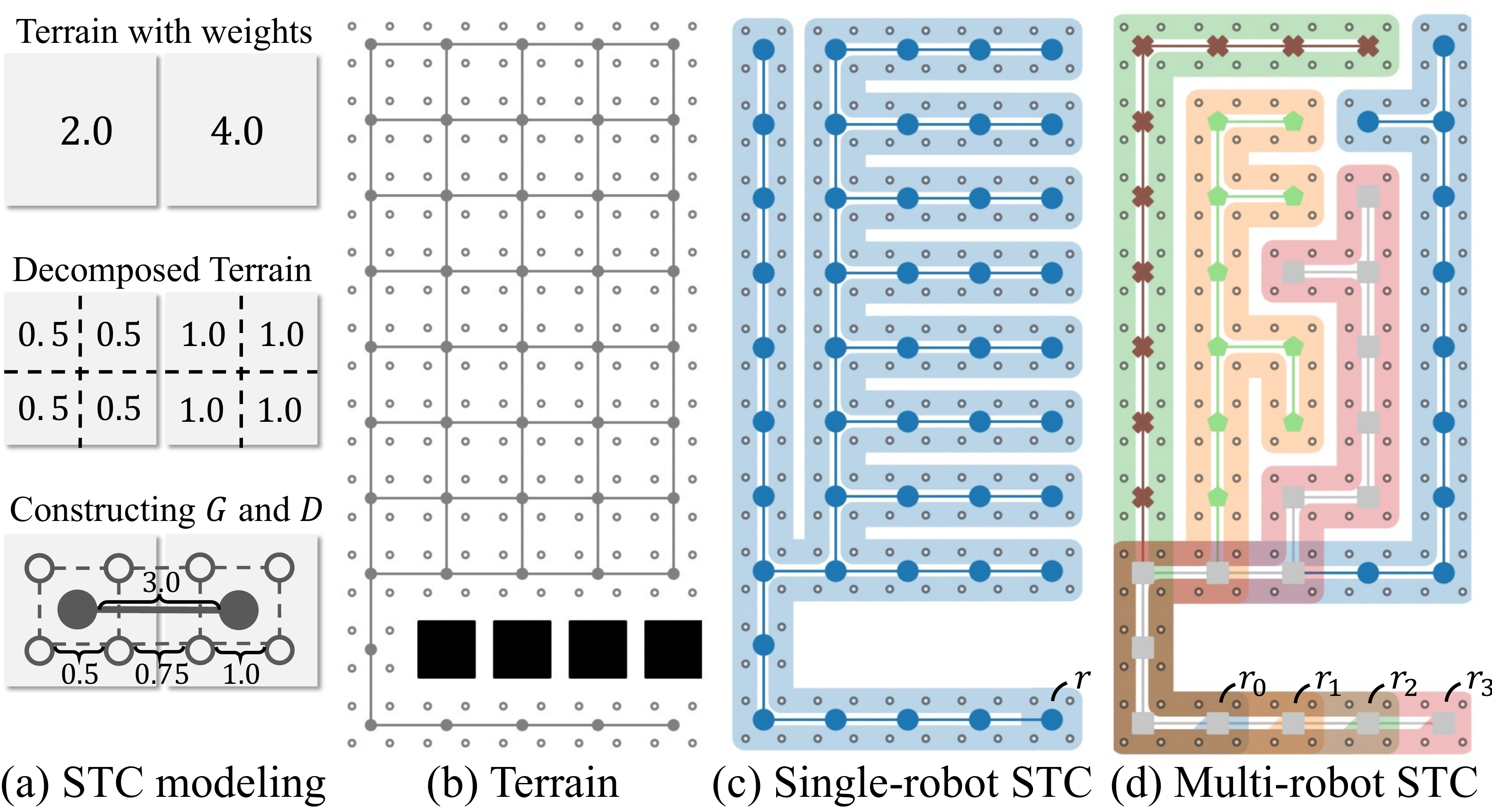}
\caption{Graph-based STC in the instance from~\cite{zheng2010multirobot, tang2021mstc}: (a) local view of STC modeling of a weighted terrain to be covered; (b) black squares, color-filled markers, and hollowed circles represent obstacles, vertices of $G$, and vertices of $D$, respectively; (c)(d) the rooted subtrees (thin lines) in $G$ and the resulting coverage paths (thick lines) in $D$.}
\label{fig:spanning_tree_coverage}
\end{figure}

We address the coverage problem for both weighted and unweighted terrains using graph-based Spanning Tree Coverage (STC) \cite{gabriely2001spanning, gabriely2002spiral}, based on the problem formulation presented in~\cite{zheng2005multi, zheng2007robot, zheng2010multirobot}. STC and its multi-robot variants operate on a 2D 4-neighbor grid graph $G=(V, E)$ that represents the terrain to be covered. Following the ``\textit{cover and return}'' setting~\cite{zheng2007robot}, the robot(s) must start and end at their respective initial vertex.
For simplicity, we restrict ourselves to a weighted terrain, where each $v\in V$ has an associated weight $w(v)$. In the case of an unweighted terrain, we assign a uniform weight of $1$ to each $v\in V$.

STC generates coverage paths on a decomposition $D=(V_d, E_d)$ of $G=(V,E)$. Each vertex $v\in V$ is decomposed into four smaller adjacent vertices, as shown in Fig.~\ref{fig:spanning_tree_coverage}-(a) and (b). Each resulting vertex $u\in V_d$ is assigned a weight $w(u)=w(v)/4$ and must be visited by at least one robot for complete coverage. The edges $(u, v)$ in both $E$ and $E_d$ are connected for each pair of vertically or horizontally adjacent vertices and are assigned a weight $\left[w(u)+w(v)\right]/2$.
The \textit{coverage time} $t(\pi)$ of a valid complete coverage path $\pi=(v_1, v_2, ..., v_{|\pi|})$ in $D$ is defined as $t(\pi)=\sum_{i=1}^{|\pi|}\left[\left(w(v_{i})+w(v_{i+1})\right)/2\right]$, where $v_1=v_{|\pi|+1}$  is the robot's given initial vertex from $V_d$.
For a CPP instance on graph $G$ and its decomposition $D$ (as depicted in Fig.~\ref{fig:spanning_tree_coverage}-(c)), STC works by circumnavigating an arbitrary spanning tree of $G$, which generates a coverage path in $D$ by always moving the robot along the right side of the spanning edges.
In practice, STC only requires the terrain graph $G$ and does not explicitly decompose it into $D$.
Since the path $\pi$ visits (i.e., enters and leaves) each vertex of $V_d$ at least once, the coverage time $t(\pi)=\sum_{i=1}^{|\pi|}w(v_{i})$ is at least the sum of the weights of all vertices in $V_d$ (which is also equal to the sum of weights of all vertices in $V$).
STC is guaranteed to generate a valid coverage path in $D$ where each $v\in V_d$ is visited exactly once (see Fig.~\ref{fig:spanning_tree_coverage}-(d)). Consequently, for CPP, STC is time-optimal using any spanning tree of $G$ in unweighted terrain~\cite{gabriely2001spanning} and time-optimal using the minimum spanning tree of $G$ in weighted terrain~\cite{zheng2007robot}.

STC has also been extended for MCPP. Formally, the objective of MCPP is to find a set $\{\pi_i\}_{i\in I}$ of $k$ paths for a given set $I=\{1, 2, ..., k\}$ of robots:
{\setlength{\abovedisplayskip}{3pt}\setlength{\belowdisplayskip}{3pt}
\begin{equation}\label{eqn:mcpp}
\{\pi_i^*\}_{i\in I}=\argmin_{\pi_1, \pi_2, ..., \pi_k}\; \max\{w(\pi_1), w(\pi_2), ..., w(\pi_k)\}
\end{equation}
}%
where $\bigcup_{i=1}^{k}\pi_i = V_d$ (all vertices in $V_d$ are covered) and each $\pi_i$ starts and ends at the initial vertex of robot $i$. STC-based approaches operate on the terrain graph $G$ and rely on a reduction from MCPP to MMRTC~\cite{even2004min, nagamochi2007approximating} that aims to find $k$ subtrees rooted at the given initial vertices of $k$ robots to cover all vertices of $V$ jointly. These approaches then use STC to convert each of the resulting $k$ subtrees into the coverage path of a robot. Lemma 4 of~\cite{zheng2010multirobot} shows that this reduction from MCPP to MMRTC achieves an asymptotic optimality ratio of 4, provided that the set of subtrees has the smallest makespan, defined as the maximum sum of edge weights of all the subtrees.
It is worth noting that solving MMRTC optimally is proven to be NP-hard for unweighted~\cite{even2004min} and weighted terrains~\cite{zheng2010multirobot}. MFC~\cite{zheng2005multi, zheng2007robot, zheng2010multirobot} adopts the RTC algorithm~\cite{even2004min} to solve MMRTC but only suboptimally. This motivates our research to develop an exact approach for MMRTC, which, to the best of our knowledge, does not exist. 
% We refer to this problem as the Min-Max Rooted Tree Cover (MMRTC) problem, which is also known as the $R$-Rooted Tree Cover problem~\cite{even2004min} or the Multi-Rooted Subtree Cover problem~\cite{nagamochi2007approximating} in the operations research literature.

\section{MIP Formulation for MMRTC}
Given an MCPP instance with the terrain graph $G=(V, E)$ and the set $I$ of robots as defined in Eqn.~(\ref{eqn:mcpp}), we define a set $R=\{r_i\}_{i\in I}\subseteq V$ of root vertices where each $r_i$ corresponds to the robot's initial vertex. The corresponding MMRTC instance aims to find a set of $k$ subtrees $\{T_{i}\}_{i\in I}$ such that each $T_i$ must be rooted at $r_i\in R$ and each vertex $v\in V$ is included in at least one subtree.
In our formulation, each edge $e\in E$ is associated with a weight $w_e$. This weight can either be computed by averaging the given weights of its two endpoints, following the standard setting from the MCPP literature as described in Sec.~\ref{subsec:stc} or be directly given by an edge-weighted terrain graph in other settings. Therefore, our formulation provides flexibility in handling both instances with weighted edges and instances with weighted vertices. Consequently, the weight of a subtree $T_i$ is defined as $w(T_i)=\sum_{e\in T_i}w_e$. The optimal set of subtrees is the one that minimizes the maximum weight among all subtrees (i.e., \textit{makespan}):
{\setlength{\abovedisplayskip}{3pt}\setlength{\belowdisplayskip}{3pt}
\begin{align}
\{T_i^*\}_{i\in I} = \argmin_{T_1,T_2,...,T_k} \max\{w(T_1), w(T_2), ..., w(T_k)\}
\end{align}
}%

To encode an MMRTC instance as a MIP model, we introduce two sets of binary variables $\textbf{x} = \{x^i_{e}\}_{e\in E}^{i\in I}$ and $ \textbf{y} = \{y^i_{v}\}_{v\in V}^{i\in I}$, where $x^i_{e}$ and $y^i_v$ take value $1$ if edge $e$ or vertex $v$ is included in the $i$-th subtree $T_i$, respectively, and $0$ otherwise.
To ensure that each subtree is a spanning tree without any cycles, we adopt an approach used in MIP models for the Steiner tree problem, as described in~\cite{cohen2019several, althaus2014algorithms}.
For this purpose, we assume that each edge has one unit of flow and introduce a set of non-negative continuous flow variables $\textbf{f} = \{f^i_{e,u}, f^i_{e,v}\}_{e\in E}^{i\in I}$ to represent the amount of flow assigned to vertices $u$ and $v$ for each edge $e=(u,v)\in E$.

Let $\tau$ denote the makespan and $e\sim v$ denote that $v$ is one of the endpoints of edge $e$. Our MIP model for MMRTC is formulated as follows:
{\setlength{\abovedisplayskip}{3pt}\setlength{\belowdisplayskip}{3pt}
\begin{align}
\label{eqn:obj}\textbf{(MIP)}&\quad\displaystyle{\minimize_{\textbf{x}, \textbf{y}, \textbf{f}, \tau}\quad\tau} &\\
\label{cstr:makespan}\text{s.t.}\quad&\sum_{e \in E} w_{e} x_{e}^i \leq \tau, &\forall i\in I \\
\label{cstr:cover}&\sum_{i\in I} y_{v}^i\geq 1, &\forall v \in V\\
\label{cstr:rooted}&y^i_{r_i} = 1, &\forall i\in I\\
\label{cstr:tree_def}&\sum_{v\in V} y_{v}^i = 1+\sum_{e\in E} x_{e}^i, &\forall i\in I\\
\label{cstr:acyclic_1} \quad\quad&f_{e,u}^i+f_{e,v}^i=x_{e}^i, &\forall e=(u,v)\in E, \forall i\in I\\
\label{cstr:acyclic_2} \quad\quad&\sum_{\substack{e\in E\\ e\sim v}}f_{e,v}^i\leq 1- \frac{1}{|V|}, &\forall v\in V, \forall i\in I
\end{align}
}%
{\setlength{\belowdisplayskip}{3pt}
\vspace{-12pt}
\begin{flalign}
\label{cstr:y_def} \hspace{26pt} x^i_{e}\leq y^i_v,  \quad\quad\quad&\forall v\in V, \forall e\in E, e\sim v, \forall i\in I
\end{flalign}
\vspace{-16pt}
\begin{flalign}
\hspace{26pt} x^i_{e}, y^i_v\in\{0, 1\},  \quad\quad\quad&\forall v\in V, \forall e\in E, \forall i\in I
\end{flalign}
\vspace{-16pt}
\begin{flalign}
\hspace{26pt} f^i_{e,u}, f^i_{e,v}, \tau\in\mathbb{R}^+, \;\;\quad&\forall e=(u, v)\in E, \forall i\in I
\end{flalign}
}%
We group the constraints of the above model as follows:
\begin{enumerate}
\item \textit{Makespan}:
Eqn.~(\ref{cstr:makespan}) ensures that $\tau$ equals the maximum weight among all the subtrees, which is minimized in the objective function defined in Eqn.~(\ref{eqn:obj});
\item \textit{Cover}:
Eqn.~(\ref{cstr:cover}) enforces that each $v\in V$ is included in at least one subtree;
\item \textit{Rooted}: 
Eqn.~(\ref{cstr:rooted}) enforces each $T_i$ is rooted at $r_i\in R$;
\item \textit{Tree}:
Eqn.~(\ref{cstr:tree_def}) ensures that each $T_i$ is either a single tree or a forest with cycles in some of its trees, while
Eqn.~(\ref{cstr:acyclic_1}) and (\ref{cstr:acyclic_2}) eliminate any cycles in $T_i$. Together, these constraints ensure that any subtree is a single tree.
\end{enumerate}

In addition to these group constraints, Eqn.~(\ref{cstr:y_def}) enforces consistency between edge variables $x^i_e$ and vertex variables $y^i_v$ for each subtree, implying that a vertex is included if and only if at least one of its incident edges is included.
While the other group constraints are straightforward to verify, the
\textit{Tree} group constraint borrowed from~\cite{cohen2019several, althaus2014algorithms} are not self-evident for our MIP model.
Therefore, we prove the following theorem, which ensures the correctness of our MIP model. With this theorem, we establish that any solution of our MIP model is a feasible solution of the corresponding MMRTC instance as defined previously.
For convenience, we will use $V(\cdot)$ and $E(\cdot)$ to denote the vertex set and edge set of a graph, respectively, in the remainder of the paper. 
\begin{theorem}
Given a solution $\{T_i\}_{i\in I}$ of the above MIP model subjected to the \textit{Tree} group constraint, every $T_i$ from the set $\{T_i\}_{i\in I}$ must be a single tree.
\end{theorem}
\begin{myproof}
For the \textit{Tree} group constraint consisting of Eqn.~(\ref{cstr:tree_def}), (\ref{cstr:acyclic_1}), and (\ref{cstr:acyclic_2}), we first denote their sub-components regarding each $i\in I$ as Eqn.~(\ref{cstr:tree_def}-$i$), (\ref{cstr:acyclic_1}-$i$), and (\ref{cstr:acyclic_2}-$i$), respectively.
Given Eqn.~(\ref{cstr:tree_def}-$i$), we can easily verify that for an arbitrary $T_i$, it is either a single tree or a forest with $C-1$ cycles in its $C$ trees.
We now prove the latter case does not hold for $T_i$ as Eqn.~(\ref{cstr:acyclic_1}-$i$) and (\ref{cstr:acyclic_2}-$i$) eliminate any potential cycles.
\begin{figure}[t]
\centering
\includegraphics[width=0.45\linewidth]{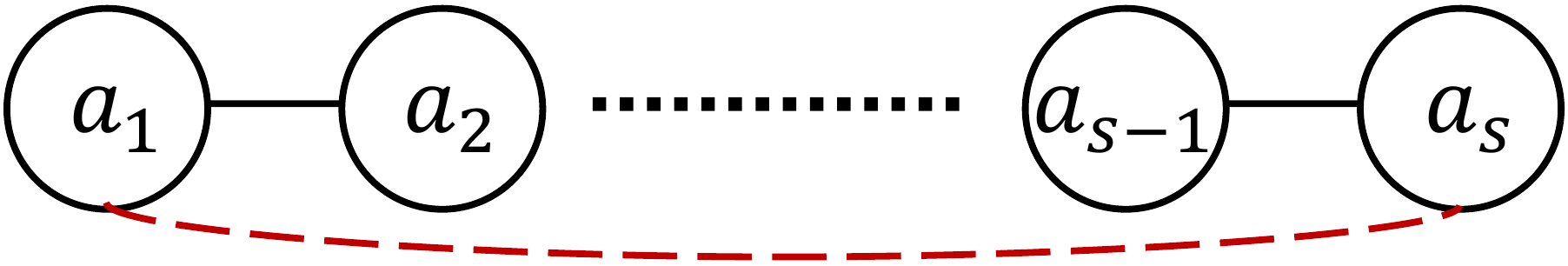}
\caption{Connected sub-component with a potential cycle.}
\label{fig:tree_proof}
\end{figure}

Given an MMRTC instance with terrain graph $G$, without loss of generality, the forest $T_i$ has an arbitrary connected component $S$ from one of its trees, where $V(S)=\{a_j\}_{ j=1,2...,s}\subsetneq V(T_i)$ and $E(S)=\{(a_j, a_{j+1})\}_{ j=1,2...,s-1}\subsetneq E(T_i)$ are included (see Fig.~\ref{fig:tree_proof}).
We assume an edge $e_c=(a_1, a_s)\in E(G)$ adds a potential cycle to $T_i$ if $e_c\in E(S)$.
By summing up Eqn.~(\ref{cstr:acyclic_1}-$i$) for all edges $e\in\{e_c\}\cup E(S)$, we have:
{\setlength{\abovedisplayskip}{3pt}\setlength{\belowdisplayskip}{3pt}
\begin{align}\label{eqn:sum_cstr_acyclic_1}
f^i_{e_c, a_1} + f^i_{e_c, a_s}+\sum_{e\in E(S)}\left(f^i_{e, u}+f^i_{e, v}\right)=x^i_{e_c} + s-1
\end{align}
}%
Denote edge set $F_i=E(T_i)/E(S)/\{e_c\}$, by summing up Eqn.~(\ref{cstr:acyclic_2}-$i$) for all vertices $v\in V(S)$, we have:
{\setlength{\abovedisplayskip}{3pt}\setlength{\belowdisplayskip}{3pt}
\begin{align*}
&\sum_{v\in V(S)}\sum_{\substack{e\in E\\e\sim v}} f^i_{e, v}\nonumber\\
=\;& f^i_{e_c, a_1} + f^i_{e_c, a_s}+\sum_{e\in E(S)}(f^i_{e, u}+f^i_{e, v})+
\sum_{v\in V(S)}\sum_{\substack{e\in F_i\\ e\sim v}} f^i_{e, v}\nonumber\\
\leq\; & s\cdot(1-\dfrac{1}{|V|})
\end{align*}
}%
combining with Eqn.~(\ref{eqn:sum_cstr_acyclic_1}), we get the following inequality:
{\setlength{\abovedisplayskip}{3pt}\setlength{\belowdisplayskip}{3pt}
\begin{align}\label{eqn:tree_proof_eq}
x^i_{e_c}\leq 1 - \dfrac{s}{|V|} - \sum_{v\in V(S)}\sum_{\substack{e\in F_i\\e\sim v}} f^i_{e, v} < 1
\end{align}
}%
thus $x^i_{e_c}\in\{0,1\}$ can only be $0$, which implies that there cannot be any cycle in $T_i$,.
Recall that $T_i$ is either a single tree or a forest with cycles, $T_i$ now must be a single tree.
Therefore, we have proved $\forall i\in I$, the subtree $T_i$ must be a single tree if the \textit{Tree} group constraint holds.
\end{myproof}

The proposed MIP model has $\mathcal{O}\left(k(|V|+3|E|)\right)$ variables and $\mathcal{O}\left(|V|+k(3+|E|+|V|)+k\bar{d}|V|\right)$ constraints, where $\bar{d}$ is the average degree of vertices depending on the structure of the MMRTC instance graph.
In this paper, all MMRTC instance graphs are 2D 4-neighbor grids; thus $\bar{d}=4$.

\section{Efficient Suboptimal Heuristics}
Although the MIP model is guaranteed to provide optimal MMRTC solutions given sufficient runtime, it becomes impractical to handle large-scale instances due to limited runtime and memory. However, we have observed that in optimal or near-optimal solutions of MMRTC instances, each subtree typically covers a neighborhood around its root, which tends to be far from the roots of other subtrees. Therefore, we propose a heuristic approach to reduce the size of the MMRTC MIP model by generating a graph $H_i$ for each subtree $T_i$ and preventing $T_i$ from covering the vertices of $H_i$. Formally, we refer to $H_i$ as the \textit{inferior graph} of $T_i$. To achieve this reduction, we replace the original terrain graph $G$ for each $T_i$ in the MIP model with a residual graph obtained by removing all the vertices and edges of $H_i$ from $G$. Consequently, all the vertex, edge, and flow variables associated with $H_i$ are removed from the MIP model for each $T_i$.

Based on the above observation, we also propose to generate a good inferior graph $H_i$ for each $T_i$ by heuristically identifying its \textit{sub-component} $H_{ij}$ with respect to each subtree $T_j$ with $j\in I/\{i\}$ such that the vertices of $H_{ij}$ are not to be included in $T_i$ since they are closer to the root $r_j$ of $T_j$ and thus inefficient to be covered by $T_i$. The inferior graph of $T_i$ is then the union of all the $k-1$ sub-components, namely $H_i=\bigcup_{j\in I/\{i\}}H_{ij}$.

%Based on this observation, we can heuristically generate a graph $H_i$ for each subtree $T_i$ to reduce the proposed MMRTC MIP model size by preventing $T_i$ from covering the vertices of $H_i$. Formally, we call $H_i$ the \textit{inferior graph} of $T_i$ where all the vertex, edge, and flow variables of $T_i$ regarding $H_i$ are removed from the MIP model in Eqn.~\ref{eqn:obj}. In other words, we replace the terrain graph $G$ for each $T_i$ in the MIP model with the residual graph obtained by removing all the vertices and edges of $H_{i}$ from $G$.
%To this end, we propose to first generate a sub-component $H_{ij}$ for each subtree $T_i$ regarding each subtree $T_j$ where $j\in I/\{i\}$, such that the vertices of $H_{ij}$ will not be in $T_i$ since they are closer to root $r_j$ thus inefficient to be covered by $T_i$. With the generated $k-1$ sub-components, the inferior graph of $T_i$ is then defined as $H_i=\bigcup_{j\in I/\{i\}}H_{ij}$.

By reducing the size of the MIP model using inferior graphs, we sacrifice optimality as the search space is reduced. However, the reduced-size MIP models should still remain complete, that is, each subtree remains a spanning tree rooted at its respective root vertex, and all subtrees jointly cover all vertices from $V$.
The following lemma shows that, if $V(H_{ij})\cap V(H_{ji})=\emptyset$, then any vertex of the inferior graph $H_i$ of subtree $T_i$ must not be in the inferior graph $H_o$ of at least one other subtree $T_o$ (and thus remain in the residual graph of $T_o$).
\begin{figure}[t]
\centering
\includegraphics[width=0.98\linewidth]{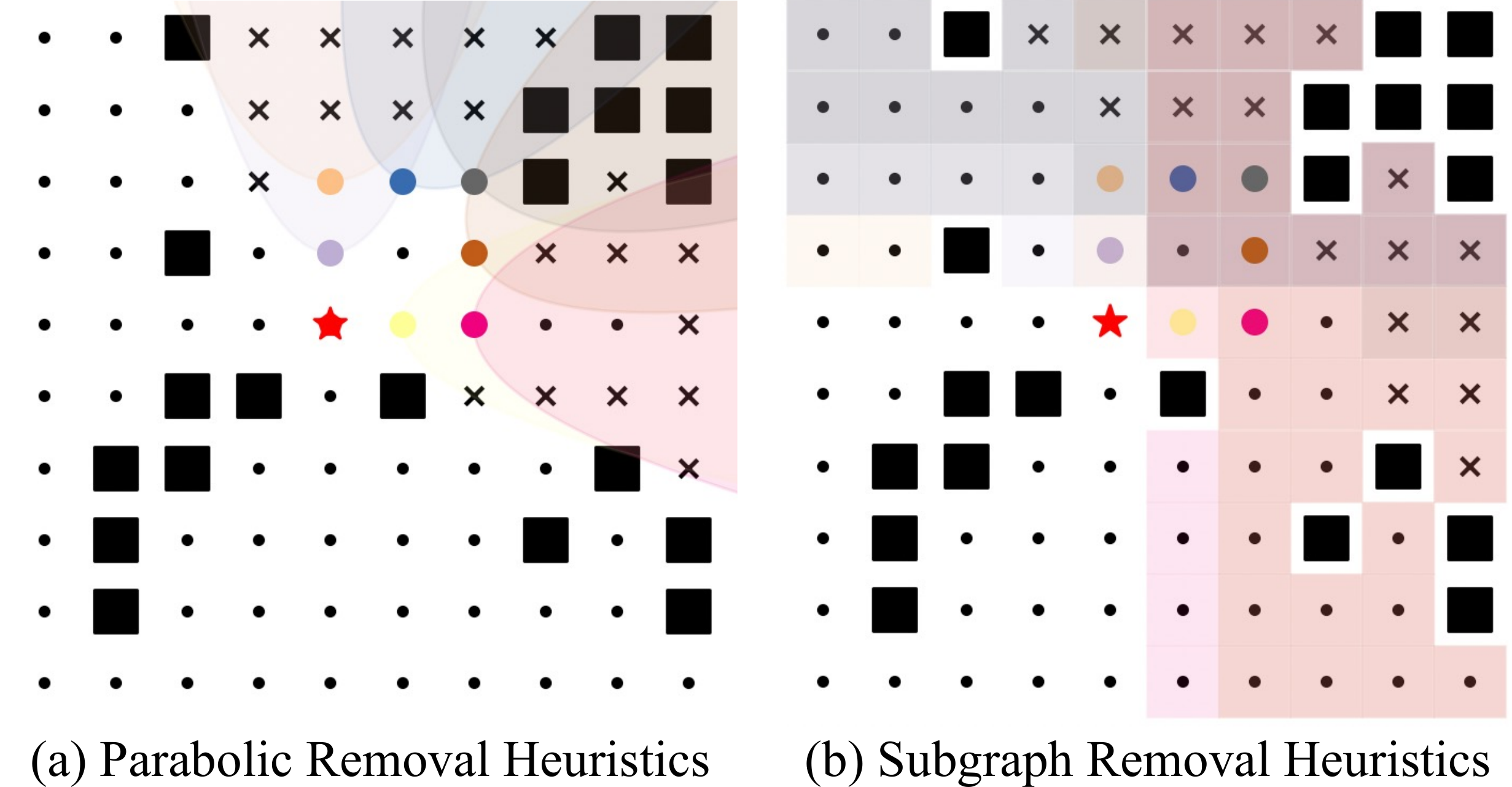}
\caption{Proposed suboptimal heuristics on the root vertex (red star) to all other root vertices (colored circles). Black squares, black crosses, and black dots are obstacles, inferior graph vertices, and residual graph vertices, respectively.}
\label{fig:heur_demo}
\end{figure}
\begin{lemma}\label{proof:completeness}\label{proof:cover}
For all $i\in I$, if $\forall j\in I/\{i\}, V(H_{ij})\cap V(H_{ji})=\emptyset$, then $\forall v\in H_i$, there exists $o\in I$ such that $v\notin V(H_o)$.
\end{lemma}
\begin{myproof}
To establish the statement $\forall v\in H_i$, there exists $o\in I$ such that $v\notin V(H_o)$, it suffices to prove that for each $j\in I/\{i\}$, $\forall v\in H_{ij}$, there exists $o\in I$ such that $v\notin V(H_o)$. 
%Consider the set $P=V(H_{ij})\cap V(H_j)$.
If $V(H_{ij})\cap V(H_j)=\emptyset$, then it follows that $\forall v\in V(H_{ij}), v\notin V(H_j)$. For the case where $V(H_{ij})\cap V(H_j)\neq\emptyset$, we can show that the statement holds as long as there exists $p\in I/\{i,j\}$ such that $Q=V(H_{ij})\cap V(H_{jp})\neq\emptyset$, given that $V(H_{ij})\cap V(H_{ji})=\emptyset$. We assume, for a proof of contradiction, that for all $p\in I/\{i,j\}$, $Q=V(H_{ij})\cap V(H_{jp})=\emptyset$. Consequently, $V(H_{ij})\cap V(H_j) = V(H_{ij})\cap \bigcup_{p\in I/\{i, j\}}V(H_{jp}) = \emptyset$, which leads to a contradiction.
%For simplicity, we assume there is only one $p\in I/\{i,j\}$ satisfying $Q\neq\emptyset$. As there is only one such $p$, we have $\forall v\in V(H_{ij})-Q, v\notin H_j$, thus we only need to prove that for arbitrary vertex $u\in Q, \exists\; q\in I/\{i,j,p\}$ such that $u\notin H_q$. Applying the same derivation above on the new set $Q$ with $V(H_{pq})\cap V(H_{qp})=\emptyset$ and so on so forth, we end up iterating over the set $I$ where we have $V(H_{ij})\cap\bigcup_{j\in I/\{j\}} V(H_j) = \emptyset$, which implies that $\forall v\in H_{ij}, \exists\;o\in I$ such that $v\notin V(H_o)$. For the cases where there are multiple $p\in I/\{i,j\}$ satisfying $Q\neq\emptyset$, the proof is similar by applying the above same derivation with multiple $Q$ sets.
\end{myproof}
Therefore, we design two heuristics in the following subsections based on the principle that $V(H_{ij})\cap V(H_{ji})=\emptyset$ to guarantee the completeness of the reduced-size models.

\subsection{Parabolic Removal Heuristic (PRH)}
For each $i\in I$ and $j\in I/\{i\}$, we define a parabola $\Omega_{ij}: y'=(a_{ij}x')^2$ under an Cartesian coordinate system $o(x', y')$, with the root vertex $r_j$ as its base and ray $r_i\rightarrow r_j$ as its symmetric axis.
The width $a_{ij}$ of $\Omega_{ij}$ is:
{\setlength{\abovedisplayskip}{3pt}\setlength{\belowdisplayskip}{3pt}
\begin{equation}\label{eqn:alpha}
a_{ij}=\alpha\cdot\sigma\left[d(r_j, c_{ij})/d({r_i, r_j})\right]
\end{equation}
}%
where $\alpha\geq 0$ is a parameter to adjust the parabola width, $d(\cdot,\cdot)$ is the shortest path distance between two vertices on $G$, and $\sigma(\cdot)$ is the logistic function that smooths the influences of the two distances.
In addition, $c_{ij}=\argmax_{v\in B}[d(r_i, v)-d(r_j, v)]$ is the vertex selected from the graph boundary vertex set $B$, which is the set of vertices whose degrees are less than 4 in a 2D 4-neighbor grid graph.
Such a design follows the aforementioned intuition in the sense that the vertices residing above $\Omega_{ij}$ are not likely to be covered by the subtree $T_i$ rooted at $r_i$, and $\Omega_{ij}$ becomes smaller when $r_i$ is close to $r_j$ or $r_j$ is close to the graph boundary.
When $\alpha=0$, $\Omega_{ij}$ is a straight line perpendicular to $r_i\rightarrow r_j$.

Let the graph induced by the inner area of $\Omega_{ij}$ be the sub-component $H_{ij}$ of $H_i$ for arbitrary $i, j\in I, i\neq j$, the Parabolic Removal Heuristic (PRH) obtains the residual graph for $T_i$ by removing the vertices and edges of $H_i$ from $G$.
Fig.~\ref{fig:heur_demo}-(a) demonstrates the relationship between the parabolas and their resulting inferior graph and residual graph with PRH.
For each subtree, the resulting residual graph might be disconnected after the removal of vertices and edges of the inferior graph.
Therefore, we adopt Alg.~\ref{alg:cc} for connectivity check to ensure the residual graph of each subtree is still connected.
We call the reduced-size MIP model after PRH and connectivity check as MIP-PRH.
\begin{algorithm}[t]
% \linespread{0.9}\selectfont
\DontPrintSemicolon
\caption{Connectivity-Check}\label{alg:cc}
\KwData{$G=(V,E)$, $r_i\in R$, $H_i$}
\KwResult{updated inferior graph $H_i$ for $T_i$}
$C_r\gets $ connected component of $r_i$ in $H_i$\;
$\mathcal{C}\gets $ list of connected components of $H_i$\;
\For{$C \in \mathcal{C}/\{C_r\}$}
{
    $v\gets $ vertex of $C$ nearest to residual graph $G-H_i$\;
    $\pi\gets $ shortest path from $v$ to $r_i$\;
    $H_i\gets (V(H_i)/V(\pi), E(H_i)/E(\pi))$ 
}
\Return $H_i$\;
\end{algorithm}

\begin{theorem}\label{proof:prh}
MIP-PRH is complete for MMRTC instances.
\end{theorem}
\begin{myproof}
We need to prove MIP-PRH produces only feasible MMRTC solutions after the vertex, edge, and flow variables corresponding to $H_{i}$ removed for each $T_i$; that is, the vertex set union from all subtrees' residual graphs equals the vertex set of the original graph and each residual graph is connected.
As we adopt Alg.~\ref{alg:cc} for connectivity check, we only need to prove that for any vertex of $H_i$, it is not included in at least one other subtree.
According to the construction rules of the parabolas in PRH, it satisfies that $\Omega_{ij}\cap\Omega_{ji}=\emptyset$ for all $\alpha\geq 0$ thus $H_{ij}\cap H_{ji}=\emptyset$ holds for arbitrary $i, j\in I, i\neq j$.
It follows that based on Lemma~\ref{proof:cover}, we have $\forall v\in H_i, \exists\;o\in I$ such that $v\notin V(H_o)$, which concludes the proof.
\end{myproof}

\subsection{Subgraph Removal Heuristic (SRH)}
Unlike the geometric approach used in PRH, the Subgraph Removal Heuristic (SRH) generates the sub-components of each inferior graph from a graph perspective as described in Alg.~\ref{alg:srh}.
Given a root $r_i\in R$ and another root $r_j\in R/\{r_i\}$, we use the Farthest-First-Search (FFS) to generate the sub-component $H_{ij}$ (see line~\ref{alg:srh-ffs} of Alg.~\ref{alg:srh}). The FFS is just a Breadth-First-Search starting from $r_i$ with the queue prioritized by the distance from each vertex to $r_i$ that ensures the vertex farthest to $r_i$ is first included in the FFS tree during the traversal, where the FFS tree size is restricted by the maximal number $b_{ij}$ of vertices given by:
{\setlength{\abovedisplayskip}{3pt}\setlength{\belowdisplayskip}{3pt}
\begin{equation}\label{eqn:beta}
b_{ij}=\left\lceil\beta\cdot|S_{ij}|\cdot\sigma\left[ d(r_i, r_j) / d(r_j, c_{ij}) \right]\right\rceil
\end{equation}
}%
where $\beta$ is a parameter to adjust $b_{ij}$, and larger $b_{ij}$ results more areas to be removed and vice versa.
$S_{ij}$ is a set of vertices defined in line~\ref{alg:srh-Sij} of Alg.~\ref{alg:srh} whose inducing subgraph the FFS will be performed on.
As we can see, the construction of the FFS tree also follows the intuition mentioned earlier, where the vertices of the FFS tree are not likely to be covered by the current subtree.
% In addition, the FFS tree size becomes smaller when $r_i$ is close to $r_j$ or $r_j$ is close to the graph boundaries.
Similar to PRH, we also need the connective check (see line~\ref{alg:srh-cc} of Alg.~\ref{alg:srh}) to ensure the connectivity of each residual graph.
Once we have the inferior graph $H_i$ returned by SRH for each $T_i$, we obtain the MIP-SRH model which only considers variables relating to the residual graph of each subtree.
Fig.~\ref{fig:heur_demo}-(b) demonstrates the inferior graph, the residual graph of one subtree with SRH, where the colored regions correspond to each $S_{ij}$.
\begin{algorithm}[t]
% \linespread{0.9}\selectfont
\DontPrintSemicolon
\caption{Subgraph-Removal-Heuristics}\label{alg:srh}
\KwData{$G=(V,E)$, $r_i\in R$, $b_{ij}$}
\KwResult{inferior graph $H_i$ of $T_i$}
$H_i\gets (\emptyset, \emptyset)$\;
$B\gets$ the set of boundary vertices of graph $G$\;
\For{$r_j \in R/\{r_i\}$}
{
    $S_{ij} \gets \{v\,|\, d(r_i, v)>d(r_j, v), v\in V\}$\;\label{alg:srh-Sij}
    $c_{ij} \gets \argmax_{v\in B}[d(r_i, v)-d(r_j, v)]$\;
    $H_{ij}\gets$ Farthest-First-Search tree rooted at $c_{ij}$ on the subgraph induced by $S_{ij}$ until $|V(H_{ij})|>b_{ij}$\;\label{alg:srh-ffs}
    $H_i\gets (V(H_i)\cup V(H_{ij}), E(H_i)\cup E(H_{ij}))$\;
}
\Return Connectivity-Check($G, r, H_i$)\;\label{alg:srh-cc}
\end{algorithm}

\begin{theorem}\label{proof:srh}
MIP-SRH is complete for MMRTC instances.
\end{theorem}
\begin{myproof}
The proof follows the same derivation in Theorem~\ref{proof:prh} for MIP-PRH.
Since the FFS (line~\ref{alg:srh-ffs}) in Alg.~\ref{alg:srh} is performed in the subgraphs induced by two disjoint vertex sets $S_{ij}$ and $S_{ji}$, we have $H_{ij}\cap H_{ji}=\emptyset$ for all $\beta \geq 0$.
Thus we have $\forall v\in H_i, \exists\;o\in I$ such that $v\notin V(H_o)$ based on Lemma~\ref{proof:cover}, which concludes the proof with the connectivity check.
\end{myproof}

\subsection{Efficiency-Optimality Tradeoff}
\begin{figure}[t]
\centering
\includegraphics[width=\linewidth]{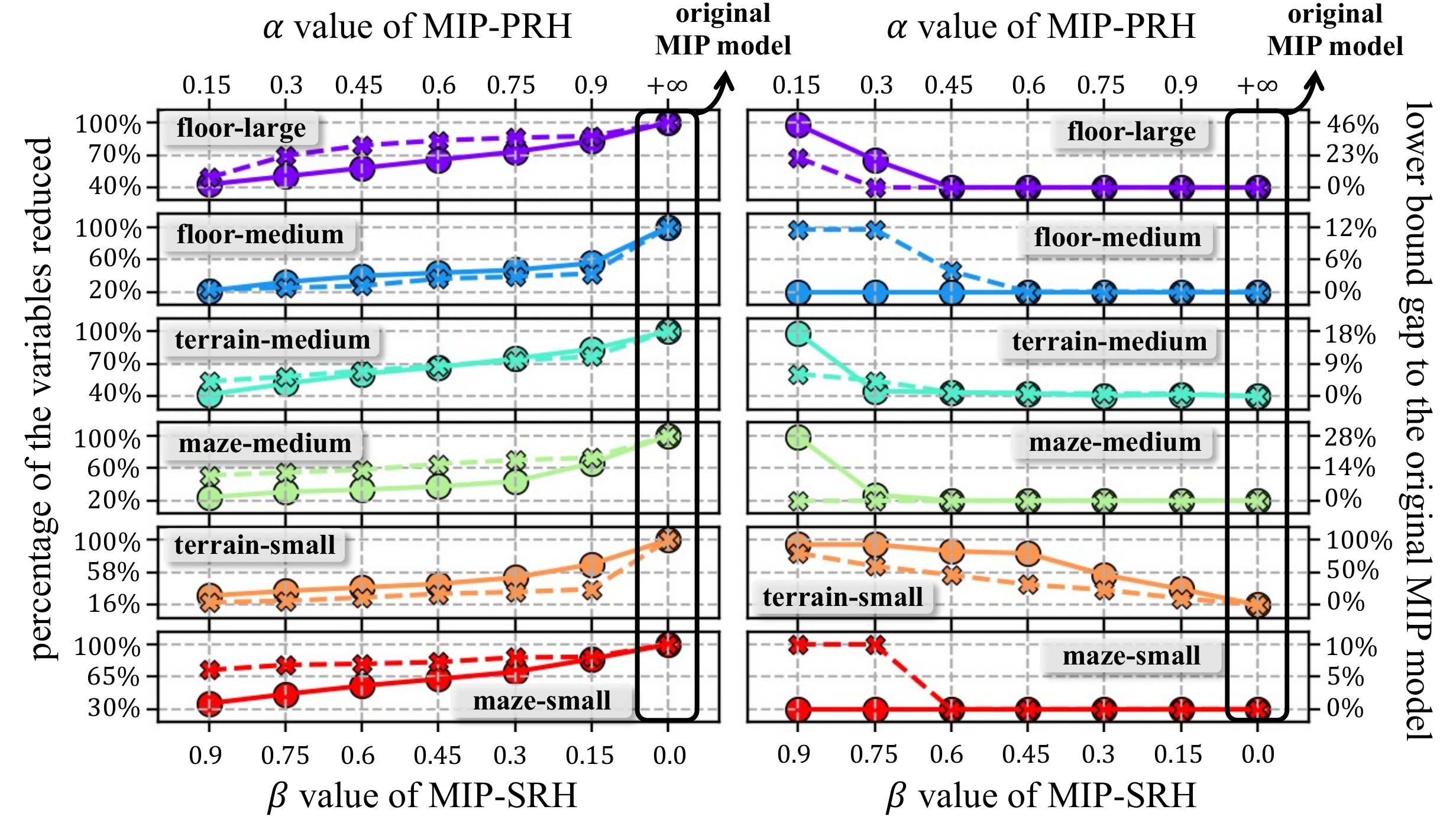}
\caption{Empirical study on MIP-PRH (crosses and dashed lines) and MIP-SRH (circles and solid lines).
Left: Percentages of the number of variables reduced from the original MIP model. Right: Gaps of the obtained lower bounds (with the 10-minute optimization) for MIP-PRH and MIP-SRH over the original MIP model.}
\label{fig:heur_exp}
\end{figure}

To verify the effectiveness of the two proposed heuristics, PRH and SRH, we conduct an empirical study on several selected MMRTC instances from Fig.~\ref{fig:istc}. The results are shown in Fig.~\ref{fig:heur_exp}, where we demonstrate the effectiveness of the variable removals and the corresponding loss of optimality for the MIP-PRH and MIP-SRH models.
We evaluate the suboptimality by comparing the lower bounds on the objective value returned by branch-and-bound based MIP solvers (e.g., Gurobi~\cite{gurobi}) within a 10-minute runtime, which are computed through the linear relaxation of the MIP models during the optimization process.
Based on our empirical findings, both removal heuristics have demonstrated their effectiveness in significantly reducing the number of variables in the MIP model. On average, PRH achieves a reduction of $42.8\%$, while SRH achieves a reduction of $50.3\%$. However, there is a tradeoff with optimality, resulting in an average loss of $2.86\%$ for PRH and $12.7\%$ for SRH. In general, SRH tends to be more versatile than PRH due to its ability to generate superior inferior graphs by considering the underlying graph structures. These observations are supported by the experimental evaluations in Sec.~\ref{sec:exp}.
To balance efficiency and optimality, we recommend selecting a smaller value of $\alpha$ for PRH or a larger value of $\beta$ for SRH when the runtime or memory is limited, or do the opposite to achieve better solution quality when optimality matters more and resources are sufficient.

%robust than PRH since SRH can better generate inferior graphs with the consideration of the graph structures, which can be observed from the experimental evaluations in Sec.~\ref{sec:exp}. In consideration of the efficiency-optimality tradeoff, it is recommended to select a smaller $\alpha$ for PRH or a larger $\beta$ for SRH if the runtime or memory is heavily limited, otherwise a larger $\alpha$ or a smaller $\beta$ is preferable to preserve the optimality.

\section{Experiments}\label{sec:exp}
\begin{figure}[t]
\centering
\includegraphics[width=0.98\linewidth]{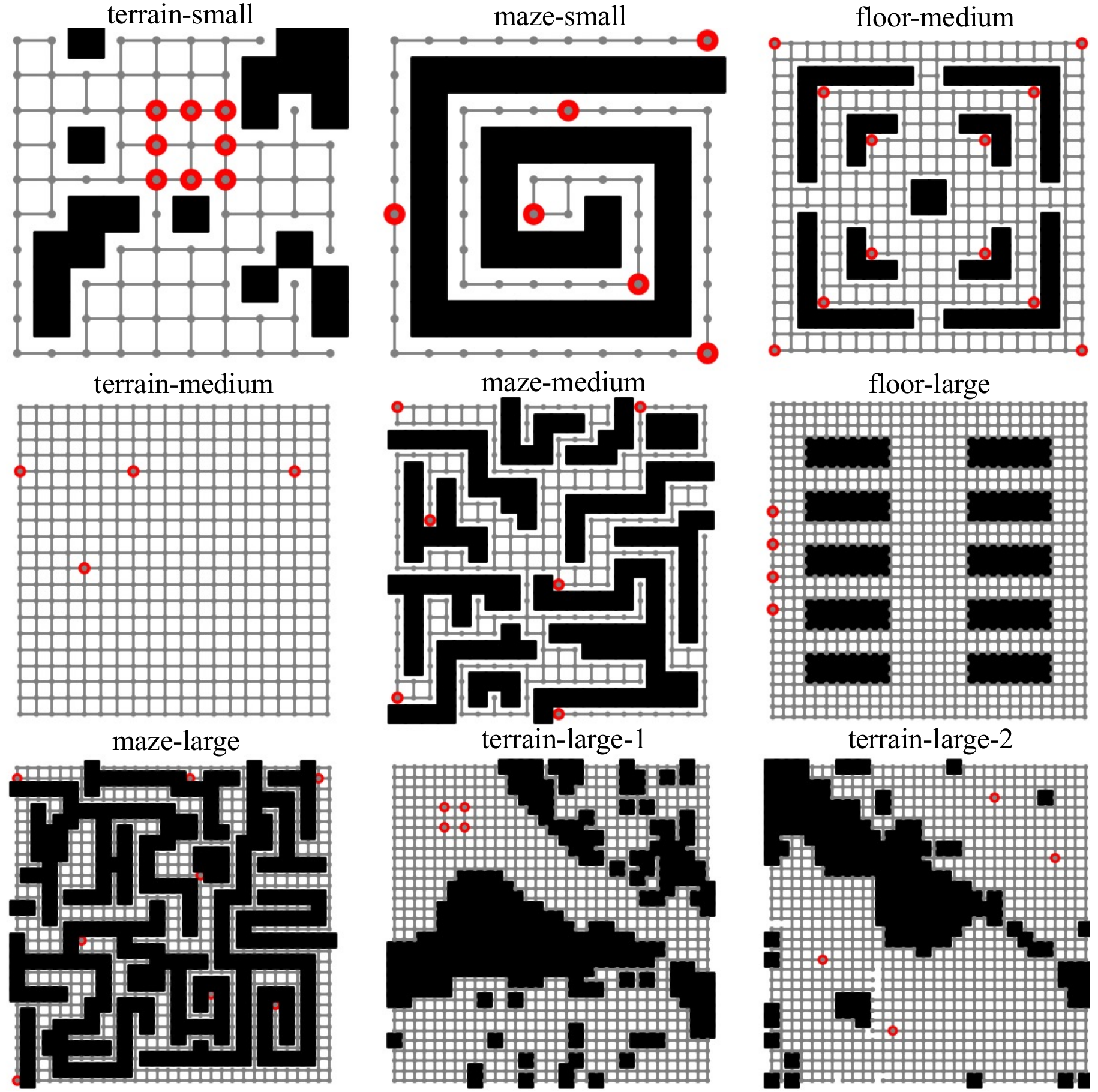}
\caption{MMRTC instances for MCPP performance evaluation. Grey circles and lines, black squares, and red circles are the terrain graphs, obstacles, and root vertices, respectively.}
\label{fig:istc}
\end{figure}

In this section, we present detailed experimental results on our proposed MIP model and its two heuristics variants, MIP-PRH and MIP-SRH, including the use of model optimization warm-startup and the performance comparisons with two state-of-the-art MCPP planners.
Tab.~\ref{tab:list_istc} specifies the details of the MMRTC instances used for MCPP performance evaluation. All instances are visualized in Fig.~\ref{fig:istc} except that \textit{floor-small} is visualized in Fig.~\ref{fig:spanning_tree_coverage}-(b).
The terrain graphs of \textit{terrain-large-1} and \textit{terrain-large-2} are generated from two large-scale outdoor satellite maps~\cite{tang2021mstc}, where the graph vertices are weighted by the estimated traversability.
For instances with weighted terrains, the weights range from 1 to 4 with a float precision of 3, whereas instances with unweighted terrains have uniform edge weights of 1.
Our implementation employs \textit{Python} to build the proposed MIP models and adopts \textit{Gurobi}~\cite{gurobi} as the MIP solver with up to 16 threads.
All experiments are executed on an \textit{Intel}\textsuperscript{\textregistered} \textit{Xeon}\textsuperscript{\textregistered} Gold 5218 CPU operating at 2.30 GHz.
Our code is open-sourced and available on \textit{Github}\footnote{\url{https://github.com/reso1/MIP-MCPP}.}.
\begin{table}[t]
\setlength\tabcolsep{1pt}
\centering
\begin{tabular}{||c|c|c|c|c|c|c|c||}
\hline
\textbf{Instance} & \textbf{Grid spec.} & \textbf{\% of obs.} & $\lvert \textbf{V} \rvert$ & 
$\lvert \textbf{E} \rvert$ & \textbf{k} & \textbf{\# of vars.} & \textbf{weighted} \\\hline
\textit{floor-small} & $5\times10$ & 8.0\% & 46 & 73 & 4 & 1061 & $\times$ \\\hline
\textit{terrain-small} & $10\times10$ & 20.0\% & 80 & 121 & 8 & 3545 & $\checkmark$ \\\hline
\textit{maze-small} & $10\times10$ & 40.0\% & 60 & 60 & 6 & 1441 & $\times$ \\\hline
\textit{floor-medium} & $20\times20$ & 19.0\% & 324 & 524 & 12 & 22753 & $\times$ \\\hline
\textit{terrain-medium} & $20\times20$ & 0.0\% & 400 & 760 & 4 & 10721 & $\checkmark$ \\\hline
\textit{maze-medium} & $20\times20$ & 39.0\% & 244 & 303 & 6 & 6919 & $\times$\\\hline
\textit{floor-large} & $30\times30$ & 15.56\% & 760 & 1370 & 4 & 19481 & $\times$\\\hline
\textit{maze-large} & $30\times30$ & 38.56\% & 553 & 717 & 8 & 21633 & $\times$\\\hline
\textit{terrain-large-1} & $32\times32$ & 27.83\% & 739 & 1275 & 4 & 18257 & $\checkmark$\\\hline
\textit{terrain-large-2} & $32\times32$ & 19.53\% & 824 & 1495 & 4 & 21237 & $\checkmark$\\\hline
\end{tabular}
\caption{Specifications for MMRTC instances. The header row corresponds to the instance name, the grid width and height, percentage of obstacles, number of vertices, number of edges, number of subtrees, number of variables, and whether the instance terrain is weighted, respectively.}
\label{tab:list_istc}
\end{table}

\subsection{Model Optimization Warm-Startup}
In this subsection, we investigate and discuss the use of the model optimization warm-startup technique in our proposed MIP models.
This technique allows the MIP solver to initialize the values of all variables with a feasible solution, typically obtained from a polynomial-time approximation algorithm.
%or solutions from another simplified optimization problem. 
This technique has been empirically shown to significantly reduce the time required to find the optimal solution. It proves advantageous and is commonly employed when solving large-size and intricate models, where finding a feasible solution can be computationally expensive.
\begin{table}[t]
\renewcommand{\arraystretch}{1.05}
\setlength\tabcolsep{3pt}
\centering
\begin{tabular}{||c|c|c|c|c|c|c|c|c|c||}
\hline
\textbf{Instance} & \textbf{Model} & \textbf{Warm} & \textbf{0.15} & \textbf{0.3} & \textbf{0.45} & \textbf{0.6} & \textbf{0.75} & \textbf{0.9} & \textbf{$+\infty$/0}\\\hline

\multirow{4}{*}{\shortstack{\textit{terrain-}\\\textit{medium}}}
& \multirow{2}{*}{\shortstack{MIP-\\PRH}} & $\times$
& 218 & 228 & 209 & 307 & 308 & / &  / \\\cline{3-10}
&  & $\checkmark$ 
& 225 & 231 & 226 & 223 & 242 & 225 & 219  \\\cline{2-10}
&\multirow{2}{*}{\shortstack{MIP-\\SRH}} & $\times$ 
& / & 215 & 209 & 209 & 209 & 239 &  / \\\cline{3-10}
&  & $\checkmark$ 
& 230 & 250 & 235 & 211 & 209 & 239 & 219 \\\hline

\multirow{4}{*}{\shortstack{\textit{maze-}\\\textit{medium}}}
& \multirow{2}{*}{\shortstack{MIP-\\PRH}} & $\times$
& / & / & / & / & / & / & / \\\cline{3-10}
&  & $\checkmark$ 
& 49 & 52 & 58 & 64 & 54 & 53 & 91   \\\cline{2-10}
&\multirow{2}{*}{\shortstack{MIP-\\SRH}} & $\times$
& / & 44 & / & 47 & 46 & 51 & /   \\\cline{3-10}
&  & $\checkmark$ 
& 61 & 46 & 48 & 47 & 47 & 51 & 91  \\\hline

\multirow{4}{*}{\shortstack{\textit{floor-}\\\textit{large}}}
& \multirow{2}{*}{\shortstack{MIP-\\PRH}} & $\times$
& 229 & / & / & / & / & / & / \\\cline{3-10}
&  & $\checkmark$ 
& 229 & 580 & 610 & 639 & 624 & 679 & 288   \\\cline{2-10}
&\multirow{2}{*}{\shortstack{MIP-\\SRH}} & $\times$
& / & / & / & / & / & 273 & /   \\\cline{3-10}
&  & $\checkmark$ 
& 665 & 595 & 374 & 350 & 226 & 273 & 288  \\\hline

\end{tabular}
\caption{Objective (i.e., $\tau$) values with and without the use of model optimization warm-startup in different MIP models with a 10-minute runtime limit. According to the header row from left to right: 1) the column \textbf{Warm} represents whether warm-startup is used; 2) the columns from 4th to 9th represent the parameters (i.e., $\alpha$ or $\beta$) used in each corresponding MIP models, and 3) the last column represents the original MIP model.}\label{tab:warmstartup}
\end{table}

%For the use of warm-startup on our models, we use the RTC~\cite{even2004min} algorithm to produce initial feasible solution for the original MIP model, and use the minimum spanning tree of each rooted subtree's residual graph as feasible solution for the MIP-PRH and MIP-SRH models.
To apply the model optimization warm-startup technique to our MIP models, we use the modified version of the RTC~\cite{even2004min} algorithm proposed in \cite{zheng2010multirobot} to produce an initial feasible solution for the original MIP model and use the minimum spanning trees on the residual graphs of the rooted subtrees as feasible solutions for the MIP-PRH and MIP-SRH models.
Tab.~\ref{tab:warmstartup} shows that this technique allows MIP, MIP-PRH, and MIP-SRH to quickly find feasible solutions for all instances within the given 10-minute runtime limit, which may not be possible without the technique. Furthermore, when the technique is applied, the resulting solutions for MIP-PRH and MIP-SRH exhibit decreased sensitivity to the values of their respective parameters (i.e., $\alpha$ and $\beta$). Overall,  it is generally advantageous to employ the technique for our proposed MIP models, even though we have observed cases where the technique yields slightly worse solutions. %it might alter the direction of solver optimization and return a slightly worse solution.
%Nevertheless, it is still recommended to equip the proposed MIP models with the warm-startup technique.

\subsection{Performance Comparison}
%Recall that similar to the MFC which extends STC, our MIP-based planner works by using STC~\cite{gabriely2001spanning} to generate coverage path on each rooted subtree obtained from the proposed MIP models. To validate its effectiveness on solving MCPP, we compare the coverage times between our MIP-based planner, MFC~\cite{zheng2010multirobot}, and MSTC$^*$~\cite{tang2021mstc}. Tab.~\ref{tab:comparison} shows the detailed performance comparison result, where the coverage time (\textbf{ct}) is computed as defined in Sec.~\ref{subsec:stc}.

In this subsection, we validate the effectiveness of our MIP-based MCPP planner by comparing it with two state-of-the-art MCPP planners, MFC~\cite{zheng2010multirobot}, and MSTC$^*$~\cite{tang2021mstc}. All these planners are based on applying STC on the terrain graph of the given MCPP instance, which ensures a fair comparison. Tab.~\ref{tab:comparison} reports the coverage times, as defined in Sec.~\ref{subsec:stc}, and runtimes.
Note that our planner uses the original MIP model only for the \textit{floor-small} instances with around 1k variables. The runtime for large-size instances is significantly long without applying PRH or SRH to restrict the number of variables in the MIP model. Therefore, for other instances, our MIP-based planner first performs two quick parameter searches from the set $\{0.3, 0.6, 0.9\}$ for the $\alpha$ parameter of MIP-PRH and the $\beta$ parameter of MIP-SRH within a short runtime (i.e., $2 \%$ of the runtime limit for each candidate), respectively, and then selects the best candidate that results in a MIP model with the smallest lower bound to solve with the remaining runtime (i.e., $88\%$ of the runtime limit). These parameter searches aim to minimize optimization loss while benefiting from the efficiency provided by PRH and SRH. All MIP models are solved with warm-startup, and their reported runtimes include the computation time of their respective warm-startup solutions.

\begin{figure}[t]
\centering
\includegraphics[width=0.98\linewidth]{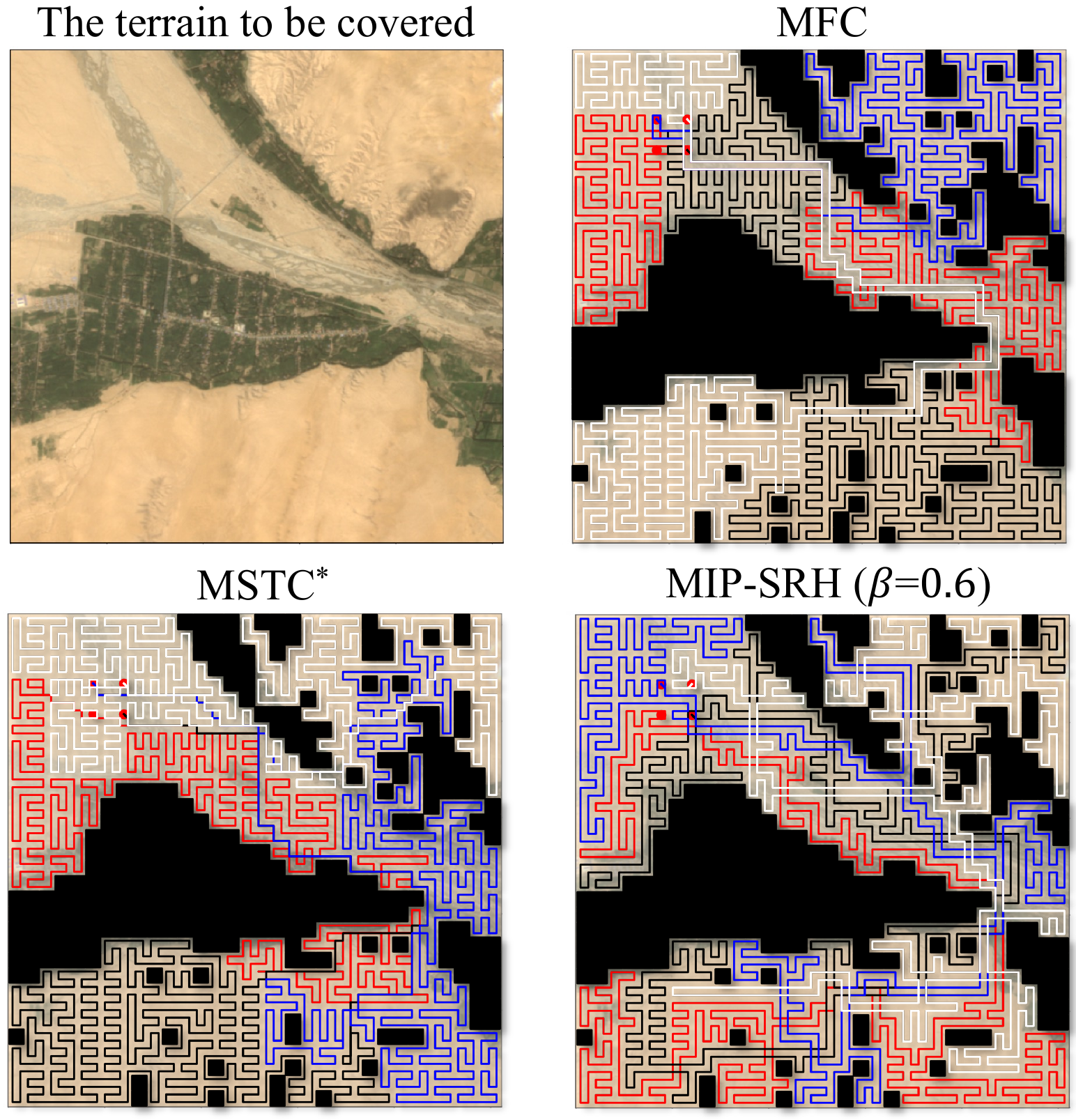}
\caption{MCPP simulation results on the \textit{terrain-large-1} instance~\cite{tang2021mstc} by MFC, MSTC$^*$ and MIP-SRH ($\beta$=$0.6$).
Lines with different colors represent coverage paths resulting from circumnavigating the rooted subtrees via STC~\cite{gabriely2001spanning}.}
\label{fig:planning_res}
\end{figure}

Overall, our MIP-based MCPP planner consistently delivers superior solution quality, resulting in an average reduction in coverage time of $27.65\%$ and $23.24\%$ compared to MFC and MSTC$^*$, respectively. The performance improvements are significant across different instance types. Specifically, the average reduction ratios compared to MFC and MSTC$^*$ are $22.61\%$ and $41.03\%$ for \textit{maze} instances, $35.65\%$ and $21.62\%$ for \textit{floor} instances, and $25.43\%$ and $11.11\%$ for \textit{terrain} instances, respectively.
Notably, our MIP-based planner performs exceptionally well for \textit{maze} instances with relatively shorter runtimes.
However, when the root vertices are clustered, particularly in a straight line, the proposed MIP models may result in a smaller reduction in coverage time. This is due to the increased likelihood of the inferior graphs of roots coinciding, requiring more time to converge.
Fig.~\ref{fig:planning_res} demonstrates the planning results of MFC, MSTC$^*$, and MIP-SRH for the \textit{terrain-large-1} instance.
% It can be observed that our planner assigns more balanced coverage workloads for robots with less overlapping.
% In addition, we report that instances with weighted edges are harder than unweighted edges, and it is even harder if the values of edge weights are of high precision.
% This is mainly due to the efforts the solver must make to improve the lower bounds during optimization.
\begin{table}[t]
\renewcommand{\arraystretch}{1.05}
\setlength\tabcolsep{1.2pt}
\centering
\begin{tabular}{||c|c|c|c||c|c|c|c||}
\hline
\textbf{Instance} & \textbf{Method} & \textbf{ct} & \textbf{rt} & \textbf{Instance} & \textbf{Method} & \textbf{ct} & \textbf{rt} \\\hline
\multirow{4}{*}{\shortstack{\textit{floor-}\\\textit{small}}} & MFC & 23.0 & 0.029 & \multirow{4}{*}{\shortstack{\textit{maze-}\\\textit{small}}} & MFC & 14.0 & 0.034 \\\cline{2-4}\cline{6-8}
& MSTC$^*$ & 21.0 & \textbf{0.016} & & MSTC$^*$ & 36.0 & \textbf{0.052} \\\cline{2-4}\cline{6-8}
& \multirow{2}{*}{\shortstack{MIP}} & \multirow{2}{*}{\shortstack{\textbf{16.0}\\(0.0\%)}} & \multirow{2}{*}{20.25} & & \multirow{2}{*}{\shortstack{MIP-PRH \\($\alpha$=$0.9$)}} & \multirow{2}{*}{\shortstack{\textbf{11.0}\\(0.0\%)}} & \multirow{2}{*}{0.064}\\
& & & & & & & \\\hline

\multirow{4}{*}{\shortstack{\textit{terrain-}\\\textit{small}}} & MFC & 36.12 & 0.073 & \multirow{4}{*}{\shortstack{\textit{terrain-}\\\textit{medium}}} & MFC & 368.2 & 0.576  \\\cline{2-4}\cline{6-8}
& MSTC$^*$ & 36.98 & \textbf{0.057} & & MSTC$^*$ & 269.5 & \textbf{0.320} \\\cline{2-4}\cline{6-8}
& \multirow{2}{*}{\shortstack{MIP-SRH\\($\beta$=$0.9$)}} & \multirow{2}{*}{\shortstack{\textbf{28.41} \\ (2.9\%)}} & \multirow{2}{*}{600} & & \multirow{2}{*}{\shortstack{MIP-SRH\\($\beta$=$0.6$)}} & \multirow{2}{*}{\shortstack{\textbf{248.8}\\(1.0\%)}} & \multirow{2}{*}{3.6e3}\\
& & & & & & & \\\hline

\multirow{4}{*}{\shortstack{\textit{floor-}\\\textit{medium}}} & MFC & 55.0 &  0.686 & \multirow{4}{*}{\shortstack{\textit{maze-}\\\textit{medium}}}
& MFC & 79.0 & 0.349 \\\cline{2-4}\cline{6-8}
& MSTC$^*$ & 47.5 & \textbf{0.174} &  & MSTC$^*$ & 65.0 & \textbf{0.340} \\\cline{2-4}\cline{6-8}
& \multirow{2}{*}{\shortstack{MIP-SRH\\($\beta$=$0.9$)}} & \multirow{2}{*}{\shortstack{\textbf{29.0}\\(7.1\%)}} & \multirow{2}{*}{3.6e3} & & \multirow{2}{*}{\shortstack{MIP-SRH \\($\beta$=$0.3$)}} & \multirow{2}{*}{\shortstack{\textbf{52.5}\\(0.0\%)}} & \multirow{2}{*}{4.936}\\
& & & & & & & \\\hline

\multirow{4}{*}{\shortstack{\textit{floor-}\\\textit{large}}} & MFC & 294.0 & 2.368 & \multirow{4}{*}{\shortstack{\textit{maze-}\\\textit{large}}}
& MFC & 105.0 & 0.997 \\\cline{2-4}\cline{6-8}
& MSTC$^*$ & 212.5 & \textbf{0.094} & & MSTC$^*$ & 139.5 & \textbf{0.575}  \\\cline{2-4}\cline{6-8}
& \multirow{2}{*}{\shortstack{MIP-PRH\\($\alpha$=$0.3$)}} & \multirow{2}{*}{\shortstack{\textbf{208.0}\\(8.7\%)}} & \multirow{2}{*}{1.08e4} & & \multirow{2}{*}{\shortstack{MIP-SRH\\($\beta$=$0.9$)}} & \multirow{2}{*}{\shortstack{\textbf{91.5}\\(0.0\%)}} & \multirow{2}{*}{54.32}\\
& & & & & & & \\\hline

\multirow{4}{*}{\shortstack{\textit{terrain-}\\\textit{large-1}}} & MFC & 597.4 & \textbf{1.672}  & \multirow{4}{*}{\shortstack{\textit{terrain-}\\\textit{large-2}}}
& MFC & 575.9 & 2.520 \\\cline{2-4}\cline{6-8}
& MSTC$^*$ & 468.6 & 1.930  & & MSTC$^*$ & 487.7 & \textbf{0.898} \\\cline{2-4}\cline{6-8}
& \multirow{2}{*}{\shortstack{MIP-PRH\\($\alpha$=$0.9$)}} & \multirow{2}{*}{\shortstack{\textbf{436.7}\\(10\%)}} & \multirow{2}{*}{2.16e4} & & \multirow{2}{*}{\shortstack{MIP-SRH\\($\beta$=$0.6$)}} & \multirow{2}{*}{\shortstack{\textbf{454.4}\\(1.2\%)}} & \multirow{2}{*}{1.08e4}\\
& & & & & & & \\\hline

\end{tabular}
\caption{Results for different planners. The columns \textbf{ct} and \textbf{rt} report the coverage time and the runtime in seconds. The percentage within each cell of the MIP-based planner reports the gap of the obtained objective (i.e., $\tau$) value of the MIP models to the obtained lower bound.}
\label{tab:comparison}
\end{table}

\section{Conclusions and Future Work}
We studied time-optimal MCPP in multi-robot systems, aiming to minimize the coverage time.
We formulated MCPP on a graph abstraction of the terrain with spanning tree coverage, where an optimal solution can be obtained with an asymptotic optimality ratio of 4 if its reduced MMRTC instance is solved to optimal. We proposed a MIP model to optimally solve the NP-hard MMRTC problem for the first time and two suboptimal heuristics to reduce the model size if given limited runtime and memory. Experimental results showed that our proposed MIP-based MCPP planner is competitive and significantly more effective than state-of-the-art MCPP planners at the cost of more runtime.

Future work can include developing specialized heuristics for instances with clustered root vertices to improve our MIP-based MCPP planner. For large-scale MMRTC instances, data-driven methods can be used to train a model to stitch or merge sub-solutions from relatively smaller instances decomposed from the original instance.
Sophisticated meta-heuristics can also be designed to select the best MIP model and parameters for different scenarios.
Furthermore, our proposed MIP model can be extended to other variants of MCPP, such as those that consider inter-robot collision avoidance, turning minimization, or robots with limited coverage capacity.

\bibliographystyle{IEEEtran}
\bibliography{ref}

\end{document}